# Staff line Removal using Generative Adversarial Networks

[a]Aishik Konwer, [a]Ayan Kumar Bhunia, [a]Abir Bhowmick, [b]Ankan Kumar Bhunia, [c]Prithaj Banerjee, [d]Partha Pratim Roy,[e]Umapada Pal

[a]Department of ECE, Institute of Engineering & Management, Kolkata, India
[b]Department of EE, Jadavpur University, Kolkata, India
[c]Department of CSE, Institute of Engineering & Management, Kolkata
[d]Department of CSE, Indian Institute of Technology Roorkee, India
[e]CVPR Unit, Indian Statistical Institute, Kolkata, India
*email: 2partharoy@gmail.com

*Abstract—* Staff line removal is a crucial pre-processing step in Optical Music Recognition. It is a challenging task to simultaneously reduce the noise and also retain the quality of music symbol context in ancient degraded music score images. In this paper we propose a novel approach for staff line removal, based on Generative Adversarial Networks. We convert staff line images into patches and feed them into a U-Net, used as Generator. The Generator intends to produce staff-less images at the output. Then the Discriminator does binary classification and differentiates between the generated fake staff-less image and real ground truth staff less image. For training, we use a Loss function which is a weighted combination of L2 loss and Adversarial loss. L2 loss minimizes the difference between real and fake staff-less image. Adversarial loss helps to retrieve more high quality textures in generated images. Thus our architecture supports solutions which are closer to ground truth and it reflects in our results. For evaluation we consider the ICDAR/GREC 2013 staff removal database. Our method achieves superior performance in comparison to other conventional approaches.

## I. INTRODUCTION

Music plays an essential role in conveying our cultural heritage. Due to various natural reasons over time, there is deterioration in the historical handwritten music documents. But it is expected to preserve this priceless information in archives. Then people will have easy accessibility to them and use it in future development purposes like retrieval, transcription etc. Though these music scores can be digitized manually, it is very time-consuming and also vulnerable to unintentional mistakes. Here comes the requirement of a robust automatic Optical Music Recognition (OMR) to retain the inherent music characteristics (tonality, notes) overcoming the mentioned limitations. OMR is difficult because in a typical music score image, there is presence of various other expressions apart from music symbols. Recent OMR approaches design machine-learning based framework.

Staff lines are a group of horizontal parallel lines, housing the music symbols. Symbols occur at different positions on staff lines, depending on sound property variations (e.g. pitch). Staff line removal is a primary module to be performed in OMR. It is done after the pre-processing stage, which uses different morphological operations. Staff lines are beneficial for correct interpretation of the musical context. But we need to separate them from the music score images for isolating the music symbols. This will result in improved and effective detection of each such symbol. The objective of staff removal is to mitigate the staff lines retaining maximum symbol information. Fig. 1 shows an ideal example of staff line removal.

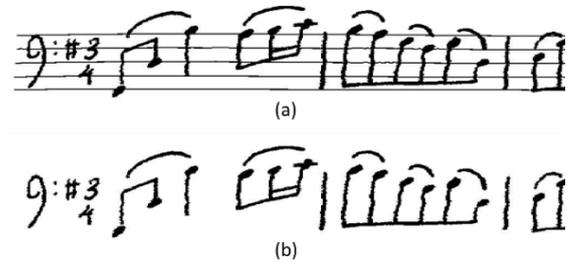

Fig.1. Example of staff line removal in a music score image

However, degrading paper quality, scanning methods and different notation styles make the process quite complex. Staff line varies for music documents over different periods. The obstacles faced in staff removal are as follows: (1) Due to shaky handwriting, lines may not be parallel. (2) Instead of being horizontal, the lines may be curved. (3) Ink distortion may lead to uneven thickness of lines. (4) Staff line color may have closer resemblance to background rather than symbols. (5) Overlapping of symbols with staff lines. Many methods have been adopted to tackle this problem. But none could succeed in performing two tasks simultaneously - reducing the noise as well as retaining the music symbol quality. The previous works bettered in either of the aspects, but only at the cost of other. We introduce a novel architecture based on Generative Adversarial Networks (GANs) that efficiently deals with the above problems.

GANs' outstanding capability to generate high quality images has been demonstrated earlier in many research works[1,2,13]. This justifies our use of Generative Adversarial training in this problem. A naive approach to deal with the staff-removal problem is to use a deep auto-encoder with pixel-wise Euclidian distance loss. But it may result in blurry images for degraded documents, and face a limitation of losing most of the music context information. Our main motivation to employ GANs in the staff-removal problem is to learn a Generator network that can generate staff-less images from the staff images. Without using a particular task-specific loss function we just specify a high-level goal for the network to produce realistic images that are indistinguishable from the real staff-less images. Blurry images will not be allowed as they are distinguishable from the actual staff-less image. To the best of our knowledge, we are the first to deal with the problem by employing GANs. In this paper we feed patches of staff-line music score images as input to a Generator network. Without using a basic encoder-decoder network as the Generator unit, we propose to use U-Net [23] with skip connections that help to improve the quality of the output image. The method in [1], demonstrates how U-Nets perform well in image translation problems. The Generator network tries to output realistic staff-less images whereas, the discriminator tries to differentiate between the generated images and the available ground truth staff-less images. Besides the GANs' adversarial loss, we use an additional L2 distance loss function to generate images that are closer to the ground truth images.

The contributions of the paper are as follows – 1) To the best of our knowledge, this is the first work which uses Generative Adversarial training for the staff removal problem. 2) We have used U-Net architecture with skip connections for our Generator network that improves the quality of the output image. 3) Our method has been tested on the ICDAR 2013 Staff Removal dataset and it achieves superior performance compared to the state-of-the-art methods.

The rest of the paper is organized as follows. In Section II we discuss some background work available for staff-removal problem and some related work in GANs. In Section III we provide the experimental setup and discuss in details the results with suitable comparisons. Finally, conclusion is given in Section IV.

## II RELATED WORK

Staff removal task is still an unsolved problem specifically when dealing with historic distorted handwritten scores. It is posed as a challenge for the document analysis community. The work in [3] showed a detailed analysis and comparisons of the available staff removal algorithms. They broadly classified the proposed methods into four divisions – 1. Skeletonization [4] which uses the property that symbols on staff lines give rise to branch points and endpoints of skeleton. 2. Vector fields [5] which associates the staff line pixels with near-zero 'Angle' values and large 'Length' values. 3. Line Tracking [6] which judges whether a staff skeleton pixel overlaps a symbol or not. 4. Run length analysis [7] based on segmentation.

However staff removal is a thought provoking problem and more recent methods have been introduced. Dutta et al.[8] treated staff line segments as parallel connections of vertical black runs with non-uniform height. Those segments which satisfied some neighboring features got approved. Su et al. [9] computes the staff line shapes after analyzing staff pixels. Then they detect the staff lines and discard them. An algorithm has been designed in [10], which considers the staff lines as maximum horizontal black paths between the score margins. The authors in [11] applied a chain of morphological operations (dilation, masking, horizontal and vertical median filtering) to directly remove staff lines from score images. Julca-Aguilar et al. [12] uses Convolutional Neural Network module on top of image operators to overcome the issue of limited input size. This entire architecture is then used for removing staff lines.

In spite of all these above mentioned algorithms, staff removal problem still contain areas for improvement. For this reason, we introduce Generative Adversarial training in our framework. Many recent methods employ GANs to achieve state-of-the-art accuracy. Our method use Adversarial loss to generate better ground truth resembling images. Thus our approach deals with different distortions (3D noise, local noise) efficiently and preserves the original music context.

Image-to-Image translation has been handled by different types of GANs. The work in [13] demonstrates the GANs ability to effectively differentiate the segmentation patches – to infer whether they are coming from segmentation module or ground truth. P. Isola and others in [1], the authors employ Generative Adversarial Networks for both input-to-output image mapping and learning loss function to train this mapping. This approach can be further applied in image synthesis, object reconstruction problems. Here they infer how Adversarial Loss is always conscious about unrealistic outputs and makes an effort for matching with original color distribution. Ledig et al. [2] present SRGAN, a Generative Adversarial Network for image super-resolution. They used a loss function which consists of an Adversarial Loss and Content Loss. The Adversarial Loss encourages the generated images solution to move closer to the original images' manifold through negative log likelihood using the discriminator which is trained to distinguish 4x super resolved images from original photorealistic ones. In [27] the authors design a novel dual-GAN mechanism, which could train the image translators from two sets of unlabeled images from two domains. In their framework, the primal GAN learns to translate domain U images to domain V images, while the dual GAN learns to invert the task. This closed loop architecture enables the images from either domain to undergo translation and reconstruction.

## III PROPOSED FRAMEWORK

In this section, we present our proposed framework. Instead of using basic pixel-wise classification approach to solve the staff-removal problem, we propose Generative Adversarial architecture. We treat the staff-removal problem as an image generation problem where we try to generate a staff removed image from staff-line image. Our model learns a mapping function from staff-line music score images to their staff-removed versions.

In this paper we intend to estimate a staff-less music score image $I^{NSL}$ from a staffline input image $I^{SL}$. Here $I^{SL}$ is the staff-line version of its staff-less ground truth $I^{GT}$. Our objective is to train a Generator function $G$ that generates a staff-less music score image from its staff-line equivalent. First, we convert the staff-line music score images into patches of size 256x256. Then we considered these patches as an input to our generator. The generator tries to generate fake staff-less image patches. Whereas a discriminator network $D$ determines how good the generator is in generating fake samples. Both the ground truth real staff-less images and the fake ones are fed into the discriminator. The discriminator predicts a probability score of the fakeness of the generated image. Both the networks are characterized by parameter $\varphi_G$ and $\varphi_D$. During training, the discriminator enforces the generator to produce realistic images of staff-less music score. Both the networks compete against each other in a min-max game. The objective function of the GAN network can be expressed as:

$$L_{GAN}(\varphi_G, \varphi_D) = \mathbb{E}_{I^{GT} \sim p_{train}(I^{GT})} \log[D_{\varphi_D}(I^{GT})] + \mathbb{E}_{I^{SL} \sim p_G(I^{SL})} \log[1 - D_{\varphi_D}(G_{\varphi_G}(I^{SL}))] \ldots (1)$$

An additional L2 distance loss function is used to force the model to generate images that are similar to the ground truth. It is beneficial to consider this traditional loss function to improve the quality of the generated images:

$$L_{L2}(\varphi_G) = \mathbb{E}_{I^{GT}, I^{SL}}\left[\left\| I^{GT} - G_{\varphi_G}(I^{SL}) \right\|_2\right] \ldots (2)$$

In the sub-section A and B we describe the detailed architecture of the generator and the discriminator network. In the sub-section C we describe the losses used in the model.

*A. Analysis of Generator Module*

Usually encoder-decoder units are used as Generator modules in problems dealing with image translation [17-20]. These units consist of a sequence of layers which perform down-sampling until a particular chokepoint layer, where the operation flips over. The disadvantage of encoder-decoder unit is that it requires all image representations to traverse each and every layer. Due to this sometimes huge amount of unwanted low-level features are exchanged between input and output, which account for the redundant information. Also, due to the down-sampling operation in the convolution layers only a part of the information of the input image is stored. It results significant information loss in each layers which cannot be used later to recover the whole image. For this purpose, we employ skip connections following the structure of a U-net [23]. We add the skip connections between layer i and layer $K - i$ where K is the total number of layers. The skip connection concatenates the layer i and layer $K - i$. It allows the network to restrict the low-level features from flowing through the net layers.

Thus, U-nets are capable in recovering images with less deterioration, which improves the overall image generation results in our case. The Generator network consists of an encoder unit with 6 convolution layers and a decoder unit with identical 6 deconvolutional layers. The filter size of each layer is 5×5 with stride step 2. Leaky ReLU and Batch Normalization are associated with feature maps of every layer. A final tanh layer is used as an activation function as it helps to stabilize the GANs' training [24]. It normalizes the output between -1 to 1. The final binary image is obtained by taking threshold value of zero.

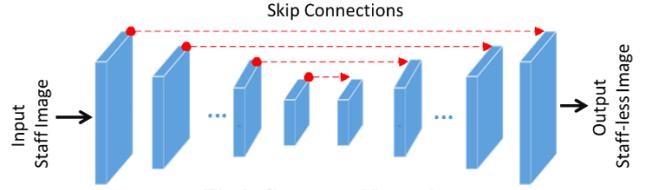

**Fig.2. Generator Network**

*B. Analysis of Discriminator Module*

For Discriminator module we define a simple Convolutional Neural Network (CNN) which has the following configuration included in the Table. The network contains 4 convolutional layers (5×5 kernel size, 2×2 stride) and 2 fully connected layers. The output is a probability of the generated image being fake.

The fundamental idea is to train the Generator in such a way that it is able to fool the discriminator network by producing photorealistic fake staff-less images which are very hard to distinguish from real ground truth images. The primary objective of the discriminator unit is to learn to differentiate the synthetically generated images from the original ones. This process makes the generated images more analogous to the ground truth images.

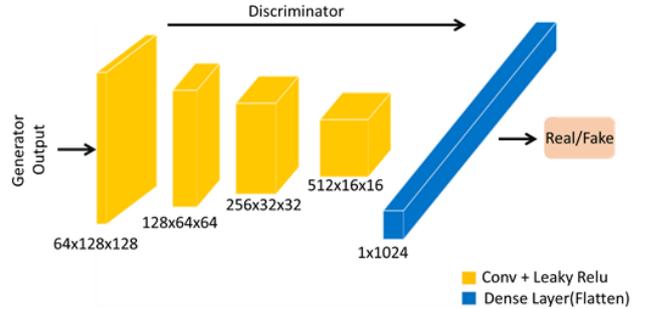

**Fig.3. Discriminator Network**

## C. Losses used in Architecture

Formulating a proper loss function is a vital step for the performance of our model. We design the required loss function $L_{net}$ as a weighted addition of GAN loss component and L2 loss component.

$$L_{net}(\varphi_G, \varphi_D) = min_{\varphi_G} max_{\varphi_D} L_{GAN}(\varphi_G, \varphi_D) + \lambda L_{L2}(\varphi_G) \quad \ldots (3)$$

The first part denotes the GAN loss. It helps our network to generate realistic staff-less images. The second part of the objective is a L2 loss that supports solutions that are closer to ground truth images. $\lambda$ is a hyper-parameter. We take $\lambda = 100$ in the experiment.

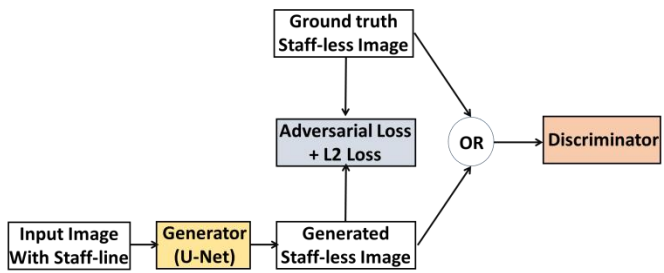

Fig.4. Proposed Architecture

## IV. EXPERIMENT

### A. Datasets

ICDAR 2013 Staff Removal dataset [21] has been used in evaluating our method. The dataset consists of 6,000 handwritten music score images of 50 musicians. Each image has a dimension of approximately 3500x2400 pixels. Images are in binary as well as grayscale format. The dataset is divided into three parts - training set (60%), validation set (10%) and testing set (30%). Images are artificially degraded adding different types of noise like Local noise, 3D noise. Thus it becomes challenging for a staff removal method to reduce the effect of this noise and preserve the music symbols in such degraded score images. A few samples of ICDAR 2013 staff-line removal datasets are shown in Fig. 5

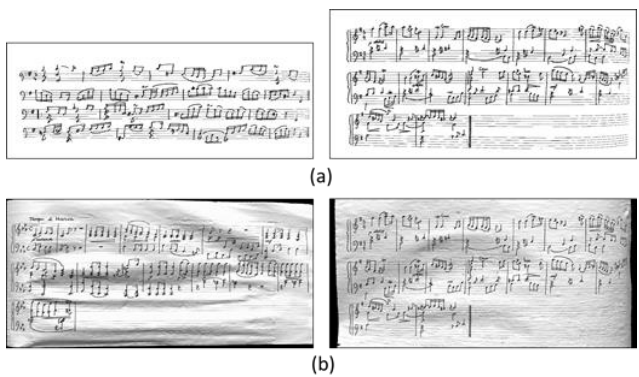

**Fig.5. Samples from ICDAR 2013 dataset (both binary and grayscale images)**

### B. Evaluation Metric

In order to evaluate the results, the F-measure (F-m) metric is considered. Let TP be the number of true positives (symbol pixels correctly classified), FP be the number of false positives (symbol pixels incorrectly classified) and FN be the number of false negatives (staff-line pixels incorrectly classified). Therefore,

$$F - m = \frac{2 \times TP}{2 \times TP + FN + FP} \quad \ldots (4)$$

### C. Baseline Method

We use a naive pixel wise classification method as a baseline approach. It is a U-net network consisting of an encoder and a decoder network. The encoder takes the image patches as input and returns a hidden feature representation of the patches. Then the decoder network outputs the staff-less image patches from the representations. We take the same architecture of the network as our Generator unit. The objective of this approach is to select those pixels that belong to a music score. Cross-entropy loss is employed to classify each pixel into two classes – foreground music symbol and background portion.

### D. Implementation Details

We normalized the image dimensions in order to extract total 32 patches of size 256x256 from each such image. Experiments are conducted on a server with Intel(R) Xeon(R) CPU E5-2670 CPU and Nvidia. We implemented the model using TensorFlow. Adam optimizer [25] with learning rate 0.001 is used to optimize the objective function. The model is trained for 60 epochs with batch size 32. Leaky ReLU (Rectified Linear Units) are used in the Convolutional layers. Batch Normalization is also applied in the architecture layers to effectively increase the training speed. During training, it is noticed that sometimes the discriminator overpowers the generator resulting in a mode collapse. Thus the training of the generator will be unstable. To overcome this problem we pre-trained the generator with only L2 distance loss for 2-3 epochs. Then the whole model is trained end-to-end. The tanh activation function of the generator outputs [24] a grayscale image where the values are normalized between -1 and 1. We take a threshold value of zero to obtain the final binary image.

### E. Comparison with state-of-the-art

We have compared our approach with state-of the-art methods like TAU, LRDE, INSEC, NUS, NUASI lin and NUASI skel, all of them participating in ICDAR 2013 staff removal competition [21]. As recent approaches (after 2016), we include Pixel, StaffNet and CNN-coupled Image operator methods.

Table I shows the results obtained by the participants in the contest for the binary case. The main idea of each method was described in Section 2 unlike TAU, which was a method specifically designed to participate in the contest. Moreover, we include the work of Calvo-Zaragoza et al. (Pixel) [22], since their results were obtained under the same conditions of the contest. In this paper, each foreground pixel is labeled as either staff or symbol. As input, a pair of scores with and without staff lines is used to train a SVM. This classification eventually results in removing staff lines. In StaffNet [26], the authors binarized the image and converted them to 23x23 patches. For each pixel, a square window patch surrounding it, is extracted from the score. A CNN is trained to distinguish between staff and symbol patches. Then the given pixel is classified as staff line or symbol depending on that patch label. Once classified, the staff pixels are gradually removed from the image. Since every time, for a single pixel, an entire patch is forwarded through the network, this method becomes computationally expensive which is a major drawback of this approach. F.D. Julca-Aguilar and others [12] addressed the issue of limiting input window size by using CNN based image operators for staff removal. However large window size may result in more misclassified pixels. Their approach is restricted to binary image and thus cannot be applied to other domains like grayscale. As can be seen, most participants are able to achieve good performance figures. However, our approach is able to improve on all the results obtained previously. This improvement is especially significant when distortions are more aggressive (for documents severely affected by Local noise and 3D noise). The comparison with Pixel method is also illustrative of the goodness of our proposal, since it demonstrates that the performance is not only achieved by using a supervised learning scheme (Pixel also does so) but because of the adequacy of the proposed model.

Table I: F-m score on binary images

| Methods | F-m score | Year |
|---|---|---|
| TAU[21] | 83.01 | 2013 |
| NUASI lin[3] | 94.29 | 2008 |
| NUASI skel[3] | 93.34 | 2008 |
| LRDE[11] | 97.14 | 2014 |
| INESC[10] | 91.01 | 2009 |
| NUS[9] | 75.24 | 2012 |
| Pixel[22] | 95.04 | 2016 |
| Image Operator [12] | 97.96 | 2017 |
| StaffNet[26] | 97.87 | 2017 |
| Baseline | 97.31 | - |
| **Our approach** | **99.32** | **-** |

Table II: F-m score on gray scale images

| Methods | F-m score | Year |
|---|---|---|
| INESC[10] | 42.09 | 2009 |
| LRDE[11] | 82.85 | 2014 |
| Pixel[22] | 90.24 | 2016 |
| StaffNet | 98.87 | 2017 |
| Baseline | 97.03 | - |
| **Our approach** | **99.14** | **-** |

We also test the method using grayscale version of the scores. Our approach can easily be used to deal with grayscale images without any additional pre-processing steps like binarization. In this case, only two of the methods submitted to the contest dealt with grayscale images: LRDE and INESC. Table II shows the results obtained by these participants, when compared to those obtained by our framework. It is clear that the participants' performance decreases remarkably, particularly as regards the INESC method. Pixel method exhibited robust characteristics and achieved promising results irrespective of the distortions. With more room for improvement, our method achieves a performance far superior to the participants. Results reported above only reflect the average performance of the methods. Our approach proves to outperform the rest of the configurations in both binary and grayscale images. We could successfully maintain a perfect balance in surpassing the noise content and sustain the music score quality for future use. This is due to our addition of combined L2 loss and Adversarial Loss and improving the reconstruction quality of generated images.

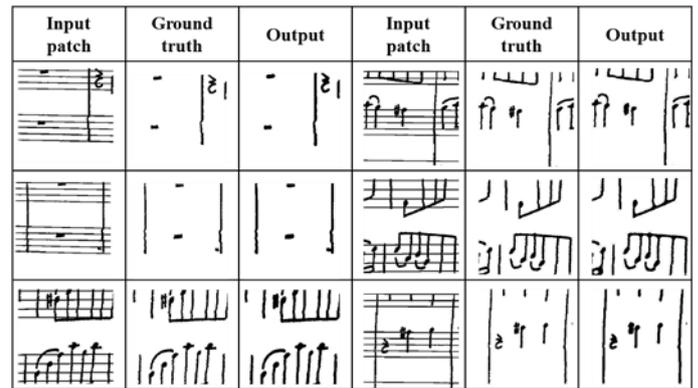

**Fig.6. Qualitative results for Binary image patches**

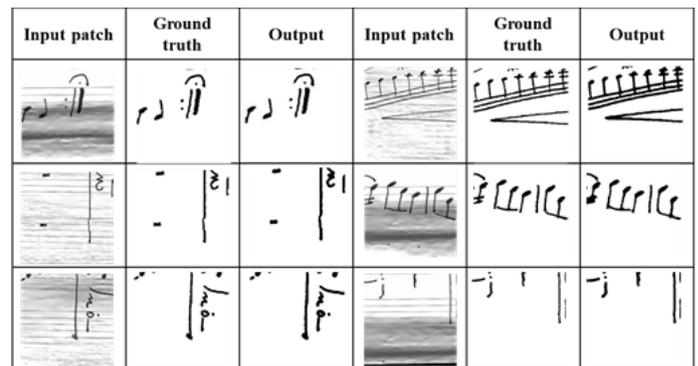

**Fig.7. Qualitative results for Grayscale image patches**

V. CONCLUSION

We proposed a novel technique for addressing Staff removal problem by using Generative Adversarial Networks. We designed the Loss function as a weighted sum of L2 loss and Adversarial loss. L2 loss helps to minimize difference between fake image and original image while Adversarial loss encourages generated images to come closer to natural manifold. We outperform naive pixel wise classification method by successful Generative Adversarial Training. Lastly our end-to-end trainable architecture, being robust to different noise, can be extended in future for ancient degraded document context analysis.

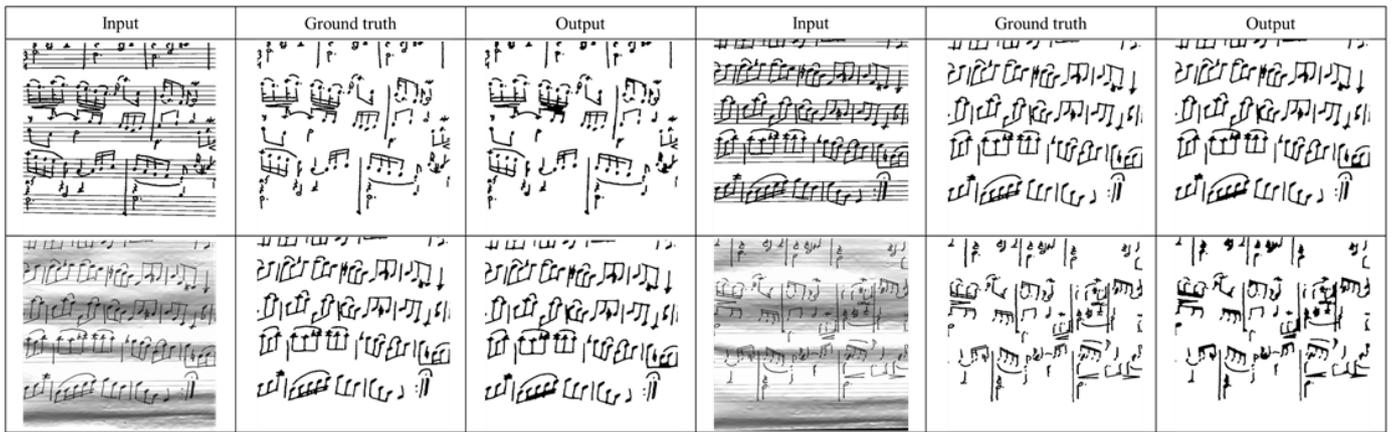

**Fig.8. Qualitative results for binary and grayscale images at page level.**